\documentclass[letterpaper, 10 pt, conference]{IEEEtran}  

\IEEEoverridecommandlockouts                              

\usepackage[utf8]{inputenc}
\usepackage[T1]{fontenc}
\usepackage{cite}
\usepackage{amsmath,amssymb,amsfonts}
\usepackage{float}
\usepackage{algorithmic}
\usepackage{graphicx}
\usepackage{textcomp}
\usepackage{xcolor}
\usepackage{etoolbox}
\patchcmd{\thebibliography}{\clubpenalty4000}{\clubpenalty10000}{}{}
\patchcmd{\thebibliography}{\widowpenalty4000}{\widowpenalty10000}{}{}
\usepackage{setspace}
\usepackage{url}
\usepackage[colorlinks=true, linkcolor=blue, citecolor=blue, urlcolor=blue]{hyperref}
\usepackage{graphics} 
\usepackage{epsfig} 
\def\BibTeX{{\rm B\kern-.05em{\sc i\kern-.025em b}\kern-.08em
    T\kern-.1667em\lower.7ex\hbox{E}\kern-.125emX}}

\begin{document}

\title{\LARGE \bf
Geometric Properties and Graph-Based Optimization of Neural Networks: Addressing Non-Linearity, Dimensionality, and Scalability
}

\author{\IEEEauthorblockN{Michael Wienczkowski} 
\IEEEauthorblockA{\textit{Department of CSE} \\
\textit{Mississippi State University}\\
Starkville, MS, USA \\
mhw205@msstate.edu}
\and
\IEEEauthorblockN{Addisu Desta} 
\IEEEauthorblockA{\textit{Department of CSE} \\
\textit{Mississippi State University}\\
Starkville, MS, USA \\
ad2398@msstate.edu}
\and
\IEEEauthorblockN{Paschal Ugochukwu} 
\IEEEauthorblockA{\textit{Department of CSE} \\
\textit{Mississippi State University}\\
Starkville, MS, USA \\
pou1@msstate.edu}
    \thanks{For any inquiries about this paper, please contact the authors at the provided emails. Code is available at \url{https://github.com/addisu-msstate/Geometric-Properties-of-NN} and additional details are in APPENDIX A.}
}

\maketitle
\thispagestyle{empty}
\pagestyle{empty}

\begin{abstract}
Deep learning models are often considered black boxes due to their large structures and complex hierarchical nonlinear transformations. To achieve the best possible predictions with limited training data, identifying the most suitable network architectures is crucial for efficiently capturing the information present in the data. Deep neural networks, characterized by many layers but relatively few connections between layers, have proven highly effective on structured data such as text, sound, and images. Understanding the geometric properties of neural networks involves examining the relationship between the architecture (network structure) and key parameters such as the number of layers, neurons, and activation functions. Geometric properties refer to the underlying structures and relationships of data manifolds, network architectures, and the transformations that occur in high-dimensional space. These factors influence how the network learns, represents data, and makes decisions in a geometric context. 

This research investigates the geometric properties and graph structures of neural networks, focusing on how these elements affect network performance and scalability. Building upon the foundational work in the seed paper\cite{c0} \textit{Graph Structure of Neural Networks}, the study explores neural networks through the lens of geometric metrics, aiming to enhance their interpretability and efficiency. The primary issue addressed is the limited understanding of the geometric structures governing neural networks, particularly the data manifolds they operate on, which complicates tasks such as classification, optimization, and representation of complex data. A key aspect of the research is examining the impact of graph structures on predictive accuracy and performance across various datasets. 

The study addresses three major challenges faced by neural networks: (1) overcoming the limitations of linear separability in handling non-linear patterns, (2) managing the dimensionality-complexity trade-off to optimize network architecture, and (3) improving scalability through advanced graph representations. By applying geometric concepts and leveraging data structure, the research proposes several strategies to address these challenges. Proposed methods include leveraging non-linear activation functions, optimizing network complexity with pruning and transfer learning, and developing efficient graph-based models. This study aims to advance the understanding of neural network geometry and structure, facilitating the development of more robust, scalable, and efficient models capable of solving complex tasks. 
\end{abstract}


\section{INTRODUCTION}
The field of Geometric Neural Networks (GNN), or understanding the geometric properties of neural networks, has grown rapidly, especially in applications that involve non-Euclidean data such as molecular modeling, social network analysis, and 3D point cloud data. Neural networks transform data through complex, multi-layered architectures, but the geometric properties underlying these transformations are not fully understood. As neural networks become more complex, there is a growing interest in understanding their geometric properties and treating them as graph-based structures. This approach aims to improve their interpretability, efficiency, and overall performance. 

In the world of AI, Deep Learning is currently the most important and widely applied technique. Despite its remarkable success, we still lack a clear, predictive understanding of how deep neural networks work and what makes them so effective at learning and generalization. A key part of this understanding comes from looking at deep neural networks through the lens of high-dimensional geometry. In particular, exploring the geometric properties of neural networks, such as their decision boundaries and the manifold structures they operate on, could provide critical insights into how these networks learn complex patterns and make generalizable predictions.

Understanding the geometric properties of neural networks requires examining the relationship between their architecture (structure) and key parameters like the number of layers, neurons, and activation functions \cite{c1}. These factors influence how the network learns, represents data, and makes decisions. Specifically, elements such as the number of layers (depth), number of neurons (width), activation functions, manifold learning \cite{c2}, optimization landscape, and generalization are all linked to the network's geometric properties. 

\section{RELATED WORK}
LeCun et al. \cite{c3} presented early work on computational graphs in neural networks, particularly in the architecture of Convolutional Neural Networks (CNNs), focusing on the flow of information through structured layers. This study laid the foundation for the graph-based analysis of neural networks by emphasizing the role of structured information flow in model performance.  

Kipf and Welling \cite{c4} introduced Graph Convolutional Networks (GCNs), expanding neural networks to process non-Euclidean data represented as graphs. GCNs are a class of neural networks that extend CNNs to graph-based data, enabling more efficient processing of non-Euclidean data. Their work demonstrated that graph structures could significantly improve the ability of models to capture relationships within complex datasets, such as social networks and molecules.  

The seed papers, You et al. \cite{c5}, \cite{c6}, \cite{c7}, \cite{c8} systematically investigates the role of graph structures in neural networks. Their analysis of graph metrics, such as clustering coefficient and average path length, identified optimal configurations that balance scalability and accuracy in models such as MLPs, CNNs, and ResNets. This study highlighted the importance of graph structure in optimizing neural architectures for predictive tasks. These papers directly address separability and geometry of object manifolds in deep neural networks and explore classification and geometry of perceptual manifolds.

Watts and Strogatz \cite{c9} introduced the small-world network model, providing key insights into the clustering coefficient and path length metrics, which are now crucial in evaluating the performance and scalability of neural networks when treated as graphs. This work forms the theoretical basis for many modern graph generation techniques applied in neural networks.  

In \cite{c10}, Chen et al. explored graph pruning and rewiring techniques as methods to enhance the efficiency of neural network architectures. By leveraging centrality measures like between-ness centrality, the study demonstrated that pruning unnecessary connections could improve both performance and computational efficiency.  

In the context of neural architecture search, Ying et al. \cite{c11} developed NAS-Bench-101, which focuses on optimizing connectivity patterns by searching for efficient Directed Acyclic Graphs (DAGs). A DAG is a graph where the edges have a direction, and no cycles are present, meaning no node can be revisited once traversed. This is relevant to the discussion on efficient graph-based network design and its implications for computational cost.  

In \cite{c12}, Barabasi and Psfai contributed foundational work in network science, which applies directly to understanding the structural properties of neural networks when modeled as graphs. Their research on scale-free networks influences how we think about network connectivity and the distribution of connections in complex architectures.  

Bassett and Bullmore \cite{c13} drew parallels between biological neural networks and artificial models, highlighting the similarities in small-world properties. This work provides insights into how the brain's efficient organization can inform the optimization of artificial neural network architectures.  

Recent work from Szegedy et al. \cite{c14} on deep convolutional networks further explores the impact of depth and complexity on scalability. Their findings are relevant to graph-based neural networks, particularly in how layer depth and graph complexity interact to influence network performance.  

Lastly, in \cite{c15}, Elsen et al. investigated sparse convolutional networks, focusing on how reducing graph complexity could enhance the computational efficiency of large-scale networks. This research aligns with the efforts to optimize graph representations for larger neural networks and complex architectures.

\section{TAXONOMY OF METHODOLOGIES}
The following taxonomy categorized the methodologies and models discussed in the literature into three main categories:  
\subsection{Models for Graph-Based Neural Networks}
\begin{itemize}
    \item Graph Convolutional Networks (GCN): Type of neural network architecture that operates on graph-structured data, extending traditional convolutional networks to non-Euclidean domains.
    \item Graph Attention Networks (GAT): Introduces attention mechanisms into graph-based learning, allowing nodes to weigh the importance of their neighbors dynamically.
    \item Message Passing Neural Networks (MPNN): Rely on the concept of message passing between nodes in a graph, where each node updates its state based on messages from its neighbors.
    \item Geometric Deep Learning (GDL): Extends deep learning to non-Euclidean domains such as graphs and manifolds, enabling the learning of representations from complex, structured data.
    \item GraphSAGE (Graph Sample and Aggregation): A scalable method for inductive representation learning on large graphs, which aggregates information from neighboring nodes to generate embeddings for graph vertices. 
    \item Deep Graph Library (DGL): A high-performance framework that facilitates the creation, training, and deployment of graph neural networks.
\end{itemize}
\subsection{Graph Structure Optimization Techniques}
\subsubsection{Graph Pruning} Process of selectively removing nodes or edges from a graph in order to simplify its structure and reduce computational costs, while attempting to maintain performance.
\begin{itemize}
    \item Centrality-Based Pruning: Involves removing nodes or edges with low centrality scores, meaning those that are less critical for the connectivity or function of the graph.
    \item Graph-Based Rewiring: Involves reorganizing or reconfiguring the connections in a graph, often to improve efficiency or reduce redundant connections.
\end{itemize}
\subsubsection{Graph-Based Sampling} Selecting a subset of nodes or edges from a graph in order to reduce its size for processing, while attempting to preserve the key structural features.
\begin{itemize}
    \item Subgraph Extraction: Involves identifying and isolating a portion of the graph (subgraph) that retains critical features, for use in a specific task or analysis.
    \item Graph Sparsification: Process of simplifying a graph by reducing the number of edges, while preserving its key properties, such as connectivity.
\end{itemize}
\subsubsection{Graph Generation Models} Methods used to create synthetic graphs with properties similar to real-world graphs.
\begin{itemize}
    \item Watts-Strogatz (WS) Model: Small-world network model that generates graphs with high clustering coefficients and short average path lengths, simulating network behavior in systems like social networks.
    \item Barabasi-Albert Model (BA): Generates scale-free networks by adding nodes that preferentially connect to existing nodes with high degrees, leading to a power-law distribution of node connections.
    \item Erdos-Renyi (ER) Model: Generates random graphs by connecting pairs of nodes with a fixed probability, leading to graphs that may lack structural patterns like small-world properties.
\end{itemize}
\subsection{Geometric Property-Based Optimization Techniques}
\subsubsection{Dimensionality Reduction Techniques} Methods used to reduce the number of variables under consideration, by transforming high-dimensional data into a lower-dimensional form, while retaining important features.
\begin{itemize}
    \item Principal Component Analysis (PCA): A linear dimensionality reduction method that transforms data into a set of orthogonal components, ordered by the variance in the data.
    \item Autoencoders: Type of neural network used for non-linear dimensionality reduction, where an encoder learns a compact representation of the input data and a decoder reconstructs the original data from this representation.
    \item Manifold Learning: Manifold Learning refers to a class of non-linear dimensionality reduction techniques that assume data lies on a low-dimensional manifold within a higher-dimensional space, such as t-SNE and Isomap.
\end{itemize}
\subsubsection{Advanced Activation Functions} Specialized functions used in neural networks to introduce non-linearity and improve the model's ability to learn complex patterns.
\begin{itemize}
    \item Polynomial Neurons: Introduces polynomial terms into the activation function, allowing neural networks to model higher-order interactions between input features.
    \item Radial Basis Functions (RBF): Type of activation function that measures the distance from a center point, often used for localized activation in high-dimensional spaces.
    \item Leaky ReLU and Parametric ReLU: Variations of the standard ReLU function, which allow small, non-zero gradients when the input is less than zero, addressing the "dying neuron" problem.
\end{itemize}

\section{CHRONOLOGICAL OVERVIEW}
This research traces significant advancements in graph-based neural networks, emphasizing the growing importance of geometric properties and graph structure optimizations. The timeline highlights key milestones, from the early exploration of computational graphs to the recent application of graph pruning and rewiring techniques. \textit{Figure 1} summarizes these key developments and progression. These developments have enhanced our understanding of how graph structures influence neural network performance, particularly in the context of scalability and predictive accuracy. Each stage reflects progress in modeling neural networks as graphs and optimizing their architectures.
\begin{enumerate}
    \item 2016: Early Exploration of Computational Graphs -
    This phase marked the use of computational graphs in traditional architectures like CNNs and MLPs, where fixed topologies constrained neural network flexibility.
    \item 2018: Introduction of Graph Convolutional Networks (GCNs) - The introduction of GCNs allowed neural networks to better handle non-Euclidean data, bringing a new level of performance in processing complex structures.
    \item 2020: Investigation of Graph Metrics - A systematic investigation into graph measures like clustering coefficient and path length revealed their critical role in optimizing neural networks, particularly in improving scalability and generalization.
    \item 2021: Graph-Based Optimization and Scaling - Practical applications of graph pruning and rewiring techniques helped improve the computational efficiency of complex neural architectures, making it feasible to apply these methods to large datasets.
\end{enumerate}  

\begin{figure}[thpb]
           \centering
           \frame{\includegraphics[width=1\linewidth]{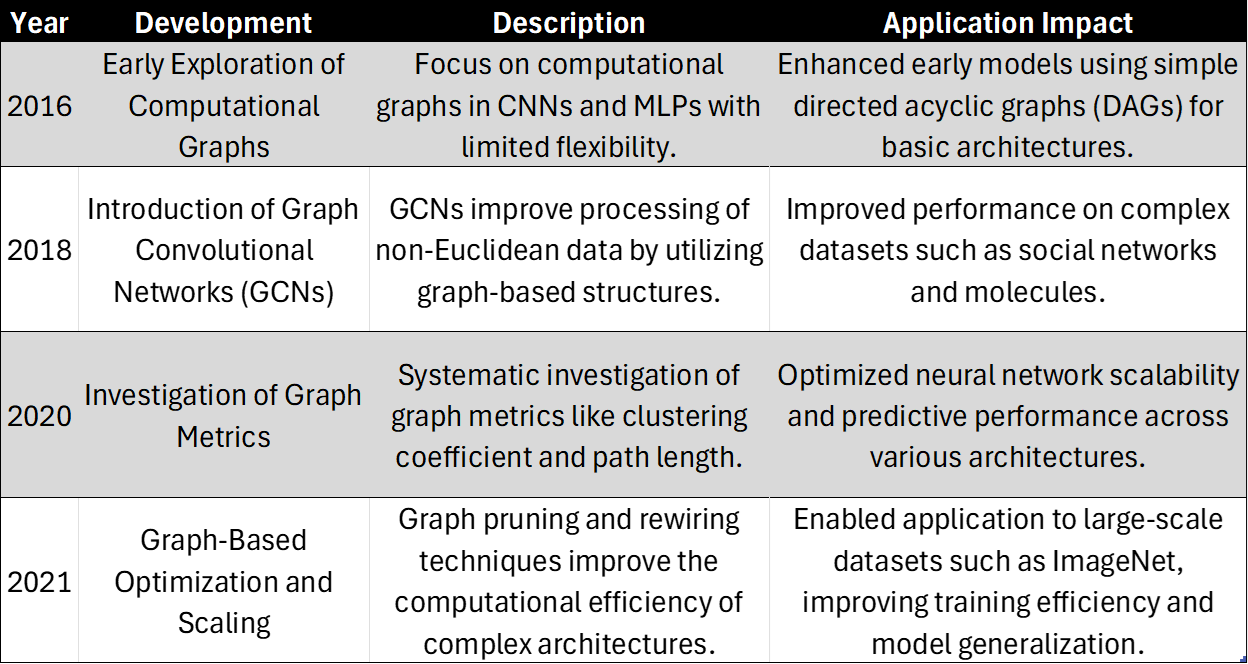}}
           \caption{Chronological Overview Table of Key Advancements in Graph-Based Neural Networks}
           \label{fig:enter-label}
       \end{figure}

\section{PROBLEM STATEMENT}
The key issue addressed in this research is the limited understanding of the geometric properties of neural networks, which affects both their interpretability and efficiency. The complexity of the network's geometry influences its learning process, impacting both optimization and generalization. This problem is significant because better geometric interpretations of neural networks can lead to improvements in various tasks, such as classification, optimization, and shape representation. A central challenge is the lack of understanding of the structure of data manifolds that influence how neural networks perform complex tasks. The geometric structures governing neural networks include the relationships between network layers, activation functions, and data manifolds, which directly impact performance in tasks like classification and optimization. The association between neural networks and geometric structures remains under-explored, and improving this understanding could result in more effective algorithms for managing complex data and optimizing performance. 

Additionally, the graph structure of neural networks plays a crucial role in their predictive performance, yet there is limited knowledge of how this structure influences accuracy. Optimizing the graph structure of neural networks could enhance their efficiency and generalizability across different datasets, which is also important for future hardware advancements. Ultimately, improving the geometric and structural comprehension of neural networks can lead to more robust and versatile models capable of performing across diverse tasks and platforms. \textit{Figure 2} summarizes the problem statement regarding the three challenges.

\begin{figure}[thpb]
           \centering
           \frame{\includegraphics[width=1\linewidth]{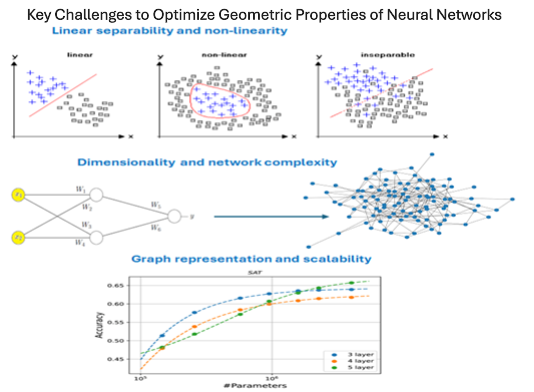}}
           \caption{Optimizing Geometric Properties of Neural Networks}
           \label{fig:enter-label}
       \end{figure}

\section{TOP THREE CHALLENGES}
\subsection{Linear Separability and Non-Linearity:}
\begin{itemize}
    \item Difficulty: Neural networks struggle with non-linearly separable data, especially in lower dimensions. Simplistic activation functions, such as step functions, are not well-suited for capturing complex patterns. Managing non-linear decision boundaries and performing geometric transformations remain difficult due to the network’s reliance on linear separability.
    \item Technical Justification: The challenge arises from the intrinsic limitations of many common activation functions, which are not capable of handling complex, non-linear patterns. Addressing this requires incorporating higher-dimensional transformations and more advanced activation mechanisms.
\end{itemize}

\subsection{Dimensionality and Network Complexity:}
\begin{itemize} 
    \item Difficulty: Neural networks need higher-dimensional transformations to classify complex data effectively. However, balancing network expressiveness (in terms of layers and neurons) with computational efficiency is difficult, especially when considering GPU limitations. More complex networks require significantly higher training costs and may lead to overfitting. 
    \item Technical Justification: The geometric properties of certain datasets demand high-dimensional transformations, which can complicate architecture design. Increased dimensionality introduces trade-offs between accuracy and computational resources, often requiring sophisticated techniques to mitigate these issues. 
\end{itemize}

\subsection{Graph Representation and Scalability:}
\begin{itemize}
    \item Difficulty: Representing neural networks as general graphs poses significant challenges, especially when trying to optimize or scale large architectures. Efficiently handling directed acyclic graphs (DAGs) and layer-wise structures becomes computationally expensive as networks grow. 
    \item Technical Justification: Scaling graph-based representations requires optimizing for graph properties like clustering coefficients and path lengths, which are computationally intensive to compute and difficult to manage for large neural networks.  
\end{itemize}

\section{PROPOSED METHODS}
\subsection{Linear Separability and Non-Linearity:}
\begin{itemize}
    \item Technique: Implement advanced activation functions and manifold transformations \cite{c16},  such as polynomial-based neurons \cite{c17}, \cite{c18}, leaky ReLU \cite{c19}, and radial basis functions (RBF) \cite{c20}, to enhance the ability of networks to manage complex, non-linear patterns. These activation functions allow for greater flexibility in decision boundary formations and better handling of intricate data. Polynomial-based neurons have demonstrated their ability to create higher-order decision boundaries suitable for non-linear separability \cite{c17}, \cite{c18}, while RBFs offer localized activations for precision modeling \cite{c20}. Leaky ReLU has proven effective in addressing the "dying neuron" problem, ensuring more thorough training \cite{c19}.
    \item Goal: Improve the network's ability to differentiate complex, non-linearly separable data by expanding the range of activation functions and transformations \cite{c21}. This aligns with previous findings that advanced activation functions significantly enhance the representational power of neural networks \cite{c19}, \cite{c20}.
\end{itemize}

\subsection{Dimensionality and Network Complexity:}
\begin{itemize} 
    \item Technique: Using optimization techniques such as pruning \cite{c22}, \cite{c23}, transfer learning \cite{c24}, and model distillation \cite{c25} to reduce the number of neurons and layers without sacrificing performance. Pruning has been shown to effectively reduce computational cost and memory usage while retaining accuracy by removing redundant parameters \cite{c22}. Transfer learning leverages pre-trained models to adapt to new tasks, reducing training time and resource requirements \cite{c24}. Model distillation, as introduced in \cite{c25}, compresses large networks into smaller, efficient models by transferring knowledge from teacher networks to student networks. By strategically simplifying the architecture, we can achieve computational efficiency while maintaining accuracy.
    \item Goal: Balance complexity and efficiency by identifying optimal network structures that maintain high performance without overburdening computational resources. This approach builds upon prior work demonstrating the trade-offs between model complexity and efficiency in constrained environments.
\end{itemize}

\subsection{Graph Representation and Scalability:}
\begin{itemize}
    \item Technique: Implement hierarchical gap encoding as the primary method to address scalability challenges in large neural networks. This technique efficiently represents large-scale graphs by encoding relationships between nodes as gaps in sorted node IDs, significantly reducing memory usage and computational costs. It also supports dynamic graph updates, such as adding or removing nodes and edges, making it suitable for scalable and adaptive applications. While gap encoding is the primary focus, complementary methods, such as Watts-Strogatz (WS) flexible models \cite{c26}, and graph-sampling techniques \cite{c27}, \cite{c28} will be reviewed for their potential to improve clustering and sparsification.
    \item Goal: Efficiently represent neural networks as graphs to facilitate better analysis and optimization, especially for larger architectures. The hierarchical gap encoding method directly addresses the memory and computational limitations of traditional adjacency matrices, enabling scaling processing of large-scale networks. This aligns with prior research \cite{c26}, \cite{c27}, \cite{c28} suggesting that graph-based representations enhance both interpretability and scalability in neural networks.
\end{itemize}


\section{METHODOLOGY}
Optimizing neural network requires addressing their inherent challenges in handling complex geometric properties, balancing dimensionality with computational efficiency, and achieving scalability in large and dynamic systems. Neural networks can be abstracted as graph-like structures, where neurons represent nodes and connections act as edges. By leveraging geometric insights and graph-based representations, this study seeks to streamline these configurations for enhanced efficiency, interpretability, and scalability.
To optimize the multifaceted challenges, we divided our approach into three distinct paths, each targeting a critical area of improvement:
\begin{itemize}
    \item Linear Separability and Non-Linearity: Focused on overcoming the limitations of simplistic activation functions by introducing advanced non-linear mechanisms, such as polynomial neurons and radial basis functions (RBF), to better handle complex decision boundaries and geometric transformations.
    \item Dimensionality and Network Complexity: Concentrated on balancing the trade-off between network expressiveness and computational efficiency through dimensionality reduction techniques (e.g., PCA \cite{c29}, autoencoders\cite{c30}) and network pruning, simplifying the geometric structure of the network without compromising accuracy.
    \item Graph Representation and Scalability: Addressed the scalability challenges of neural networks by employing hierarchical gap encoding to optimize graph properties, such as clustering coefficients and path lengths. This approach enables efficient representation, dynamic updates, and performance scaling for large or complex graphs.
\end{itemize}
By dividing the research into these three complementary paths, we ensured a comprehensive exploration of neural network optimization through geometric properties and graph-based methodologies.

\subsection{Linear Separability and Non-Linearity}
To address the challenges of linear separability in high-dimensional data, our methodology incorporates advanced activation functions that allow neural networks to model complex, non-linear decision boundaries. Linear separability represents a fundamental limitation for many neural networks, particularly when processing complex, non-linear patterns in data. Neural activity, whether biological or artificial, can be visualized as points in high-dimensional spaces structured within geometric constructs called manifolds. These manifolds provide critical insights into how neural systems transform non-linearly separable data into structured, linearly separable forms through geometric modifications across network layers. This geometric perspective highlights the importance of designing advanced activation mechanisms that can effectively manipulate such transformations.

Traditional activation functions, such as step functions and standard ReLU, are inherently limited in their ability to model intricate decision boundaries. These functions operate under assumptions of linear transformations, which are inadequate for capturing the complex geometric relationships present in many real-world datasets. Overcoming these limitations requires advanced activation mechanisms capable of providing greater flexibility in modeling decision boundaries and geometric transformations.

Our approach focused on introducing polynomial neurons, radial basis functions (RBFs), and leaky ReLU variants to expand the range of non-linear transformations available to neural networks. These advanced mechanisms allow networks to better handle complex datasets, improving their representational power and generalization capabilities.

\subsubsection{Polynomial Neurons for Higher-Order Decision Boundaries}
Polynomial neurons extend traditional activation functions by incorporating polynomial transformations, enabling the network to model higher-order interactions between input features. In neural networks, hidden layers implicitly perform manifold transformations \cite{c30}, progressively flattening or unwrapping complex data structures to enhance linear separability. Polynomial neurons take this a step further by reshaping complex, highly curved manifolds into more structured forms that are amenable to linear linear decision boundaries. This untangling of manifold geometry allows for better representation of intricate patterns, particularly in datasets with significant non-linear relationships. Unlike standard functions, which rely on linear approximations, polynomial neurons can create intricate, curved decision boundaries that are more representative of non-linear data distributions. 

\noindent The equation for a polynomial neuron is as follows:
\[
y = \sum_{i=1}^n w_i x_i^d + b
\]
where $d$ represents the degree of the polynomial, $x_i$ are the input features, $w_i$ are weights, and $b$ is the bias. The degree $d$ can be adjusted to adapt to the complexity of the task, offering flexibility in modeling intricate relationships. 

\noindent Algorithmically, polynomial neurons operate by:

\begin{enumerate}
    \item Applying polynomial transformations to the input data.
    \item Passing the transformed data through subsequent layers for further learning.
    \item Utilizing back propagation to iteratively update weights and biases, ensuring that the decision boundaries align with the underlying data distribution.
\end{enumerate}

\noindent For example, datasets such as XOR and concentric circles, which are inherently non-linearly separable, demonstrated the shortcomings of traditional activations but were effectively modeled by polynomial neurons. This approach enables the network to create non-linear decision boundaries that would be difficult for traditional activation functions to capture, particularly in cases like XOR or concentric circles, where data is inherently non-linearly separable.

The flexibility of polynomial neurons lies in their ability to adapt the degree of the polynomial to the complexity of the task. For instance, low-degree polynomials may suffice for moderately complex patterns, while higher-degree polynomials provide greater flexibility for more intricate relationships. This adaptability allows for a balance between complexity and computational efficiency. However, as higher-degree polynomials risk overfitting, regularization techniques such as weight decay were incorporated to constrain their expressive power and maintain generalization.

\subsubsection{Radial Basis Functions for Localized Representations}
Radial Basis Functions (RBFs) provide a localized activation mechanism that enhances the network's precision in modeling specific regions of the input space. Localized activation through RBFs modifies the curvature of neural manifolds, focusing on specific regions of the input space to reduce the complexity of the overall structure and improve separability. 

\noindent The RBF activation is defined as:
\[
\phi(x) = \exp\left(-\frac{\|x - c\|^2}{2\sigma^2}\right)
\]

\noindent where $c$ represents the center of the RBF, $\sigma$ controls the spread (or width) of the function, and $||x - c||$ is the Euclidean distance between the input $x$ and the center. In comparison to 

\noindent Algorithm for RBF Implementation:
\begin{enumerate}
    \item Initialize RBF centers $c$ and spreads $\sigma$
    \item Compute the activation $\phi(x)$ for each input.
    \item Aggregate the outputs of the RBF units for classification or regression.
\end{enumerate}

\noindent In comparison to polynomial neurons, RBFs focus on local regions of the data, making them more efficient for tasks where local patterns dominate, such as in clustering tasks like the Iris dataset.

By targeting high-density regions of the manifold, RBFs simplify data distributions and ensure that local clusters are accurately classified, even when global data structures are complex or noisy. Unlike global functions, RBFs focus their influence within a localized area, reducing the impact of noise and irrelevant features. This localized activation is particularly effective in high-dimensional feature spaces, where the data often forms clusters or regions of importance that require precise modeling.

The RBF implementation was evaluated on datasets with clearly defined clusters, such as the Iris dataset, as well as synthetic data like Gaussian mixtures. By placing RBF centers strategically, the network achieved accurate decision boundaries that aligned with the natural geometry of the data. Furthermore, the compact activation regions of RBFs reduced overfitting risks, particularly in small or sparse datasets, while maintaining high accuracy in classification tasks.

\subsubsection{Leaky ReLU Variants for Robust Training}
Traditional ReLU functions, while popular for their simplicity and efficiency, suffer from a significant drawback: the "dying neuron" problem. This occurs when neurons become inactive due to zero gradients, leading to stagnation in learning. To address this, we incorporated leaky and parametric ReLU variants, which allow small gradients in negative activation regions. 

\noindent The Leaky ReLU function is defined as:
\[
f(x) =
\begin{cases}
x & \text{if } x > 0, \\
\alpha x & \text{if } x \leq 0
\end{cases}
\]

\noindent where $\alpha$ is a small constant (e.g., $\alpha$ = 0.01) that controls the slope for negative inputs. This ensures that neurons in negative activation range contribute to gradient updates.

\noindent Algorithm for implementing Leaky ReLU:
\begin{enumerate}
    \item Apply the Leaky ReLU function to the input $x$.
    \item Compute gradients for all inputs during backpropagation, including those in the negative range.
    \item Update weights and biases to ensure consistent learning across all neurons.
\end{enumerate}

Leaky ReLU's ability to prevent neurons from becoming inactive allows it to maintain stable learning during training, particularly in deeper architectures. These improvements were evident in experiments on the MNIST dataset, where leaky ReLU variants outperformed traditional ReLU by achieving faster convergence and higher accuracy. By introducing a small slope for negative inputs, these variants prevent neurons from becoming permanently inactive, ensuring consistent weight updates during training.

In addition to mitigating the dying neuron problem, these variants improved stability in training deeper networks. For example, parametric ReLU (PReLU), with its learnable slope parameter, further enhanced the network's adaptability, particularly in tasks involving imbalanced or noisy data. These improvements were evident in experiments on the MNIST dataset, where leaky ReLU variants outperformed traditional ReLU by achieving faster convergence and higher accuracy.

\subsubsection{Performance and Evaluation}
To evaluate the effectiveness of these advanced activation mechanisms, we conducted extensive experiments on synthetic datasets, such as XOR and concentric circles, as well as real-world benchmarkls like MNIST and Fashion-MNIST. These datasets offered varying levels of complexity, allowing us to assess the performance improvements provided by each mechanism. For the XOR dataset, polynomial neurons demonstrated their ability to handle non-linear separability by creating intricate decision boundaries, achieving near-perfect accuracy. On the MNIST dataset, RBFs and leaky ReLU variants showed significant improvements in convergence speed and robustness compared to traditional activations.

Metrics such as classification accuracy, convergence time, and decision boundary complexity were used to evaluate the methods. Additionally, visualizations of decision boundaries provided qualitative insights into how these advanced activation functions transformed the network's representational capacity. For example, polynomial neurons produced smooth, curved boundaries that aligned closely with the true data distribution, while RBFs generated localized boundaries that precisely captured cluster geometries. The introduction of advanced activation functions, including polynomial neurons, RBFs, and leaky ReLU variants, significantly enhances the ability of neural networks to handle non-linear separability. 

\subsection{Dimensionality and Network Complexity}
The trade-off between dimensionality and network complexity is a fundamental challenge in neural network design, particularly when viewed through the lens of their geometric properties. Neural networks can be conceptualized as graph-like structures, where neurons represent nodes and edges. Managing the complexity of these geometric configurations is essential for ensuring computational efficiency and generalizability, especially in resource-constrained environments. This challenge becomes even more pronounced in real-world applications, where the availability of computational resources varies significantly. Large-scale neural networks, while powerful, often require optimization techniques to function effectively in constrained environments such as embedded systems or mobile devices.

By optimizing the geometric and graph-based representations of neural networks, this section aims to reduce the unnecessary complexity, enhance computational efficiency, and improve scalability without sacrificing performance. Our methodology incorporates dimensionality reduction, network pruning, and graph-inspired optimization techniques to better understand and streamline the geometric properties of neural networks.  

\subsubsection{Dimensionality Reduction as a Geometric Simplification}
Dimensionality reduction techniques were implemented to simplify the input feature space and reduce the dimensionality of intermediate feature maps. This aligns with the geometric interpretation of neural networks, where high-dimensional transformations can be visualized as mappings across a manifold. These mappings provide valuable insights into the structure of the data, revealing intrinsic properties such as clusters or separable regions that can simplify downstream processing. By reducing dimensionality, we effectively reduce noise and focus computational resources on the most informative features.

Principal Component Analysis (PCA) was employed to identify and train the most significant geometric components of the feature space. PCA is a linear technique for reducing the dimensionality of data by projecting it onto a lower-dimensional subspace that retains the maximum variance. The method involves finding the principal components, which are the directions of maximum variance in the data.

\noindent Principal Component Analysis (PCA) can be broken down as follows:

\noindent Covariance matrix equation:
\[
\mathbf{C} = \frac{1}{n} \mathbf{X}^\top \mathbf{X}
\]

\noindent Eigenvector equation for principal components:
\[
\mathbf{C} \mathbf{v}_i = \lambda_i \mathbf{v}_i
\]

\noindent Projection equation:
\[
\mathbf{Z} = \mathbf{X} \mathbf{W}, \quad \mathbf{W} = [\mathbf{v}_1, \mathbf{v}_2, \ldots, \mathbf{v}_k]
\]

\noindent Algorithmically, PCA operates as follows:
\begin{enumerate}
    \item Compute the covariance matrix \( \mathbf{C} \) of the dataset.
    \item Find the eigenvectors \( \mathbf{v}_i \) and eigenvalues \( \lambda_i \) of \( \mathbf{C} \).
    \item Select the top \( k \) eigenvectors based on the largest eigenvalues.
    \item Project the dataset onto the subspace spanned by these eigenvectors.
\end{enumerate}

By projecting the data onto a lower-dimensional subspace, we preserved critical information while reducing the computational cost of training downstream layers. This can be seen as collapsing the feature manifold into a more compact representation. 

Autoencoders provided a non-linear geometric transformation of the input data. Autoencoders perform non-linear dimensionality reduction by learning a compact, latent representation of the data through a neural network. The encoder maps the input \( \mathbf{x} \) to a latent space \( \mathbf{z} \), while the decoder reconstructs the input from \( \mathbf{z} \).

\noindent Autoencoders can be broken down with the following equations:

\noindent Encoder and decoder functions:
\[
\mathbf{z} = f_\theta(\mathbf{x}), \quad \hat{\mathbf{x}} = g_\phi(\mathbf{z})
\]

\noindent Reconstruction loss equation:
\[
\mathcal{L}_{\text{reconstruction}} = \frac{1}{n} \sum_{i=1}^n \| \mathbf{x}_i - \hat{\mathbf{x}}_i \|^2
\]

\noindent Algorithmically, autoencoders operate as follows:
\begin{enumerate}
    \item Train the encoder \( f_\theta \) and decoder \( g_\phi \) on the dataset by minimizing the reconstruction loss \( \mathcal{L}_{\text{reconstruction}} \).
    \item Use the encoder \( f_\theta \) to map input data \( \mathbf{x} \) into the latent space \( \mathbf{z} \).
    \item Use the decoder \( g_\phi \) to reconstruct \( \mathbf{x} \) from \( \mathbf{z} \) during training.
    \item Extract the latent representations \( \mathbf{z} \) for use in downstream tasks.
\end{enumerate}

Unlike PCA, autoencoders allowed us to capture the non-linear structure of the data manifold, creating a latent representation that was more compact yet expressive. This latent space was fed into subsequent layers, reducing the geometric complexity of the network. In addition to dimensionality reduction, autoencoders enabled feature extraction by learning latent representations that emphasized the most discriminative characteristics of the data. These representations proved especially useful in tasks requiring compact and generalizable feature sets, such as image classification and anomaly detection.

\subsubsection{Network Pruning as Graph Optimization}
Pruning was applied as a graph-based optimization technique, where the neural network was treated as a directed acyclic graph (DAG). Each neuron and connection was evaluated for its contribution to the overall structure and performance of the network. \textit{Figure 3} illustrates the structural transformation of a neural network before and after pruning. Redundant nodes and connections are removed, leading to a more compact and computationally efficient architecture. This evaluation involved metrics such as activation frequency and weight magnitude, which provided quantitative measures of a node's relevance. By prioritizing the retention of high-contributing neurons, the pruning process preserved critical pathways within the network.

\begin{figure}[thpb]
           \centering
           \frame{\includegraphics[width=1\linewidth]{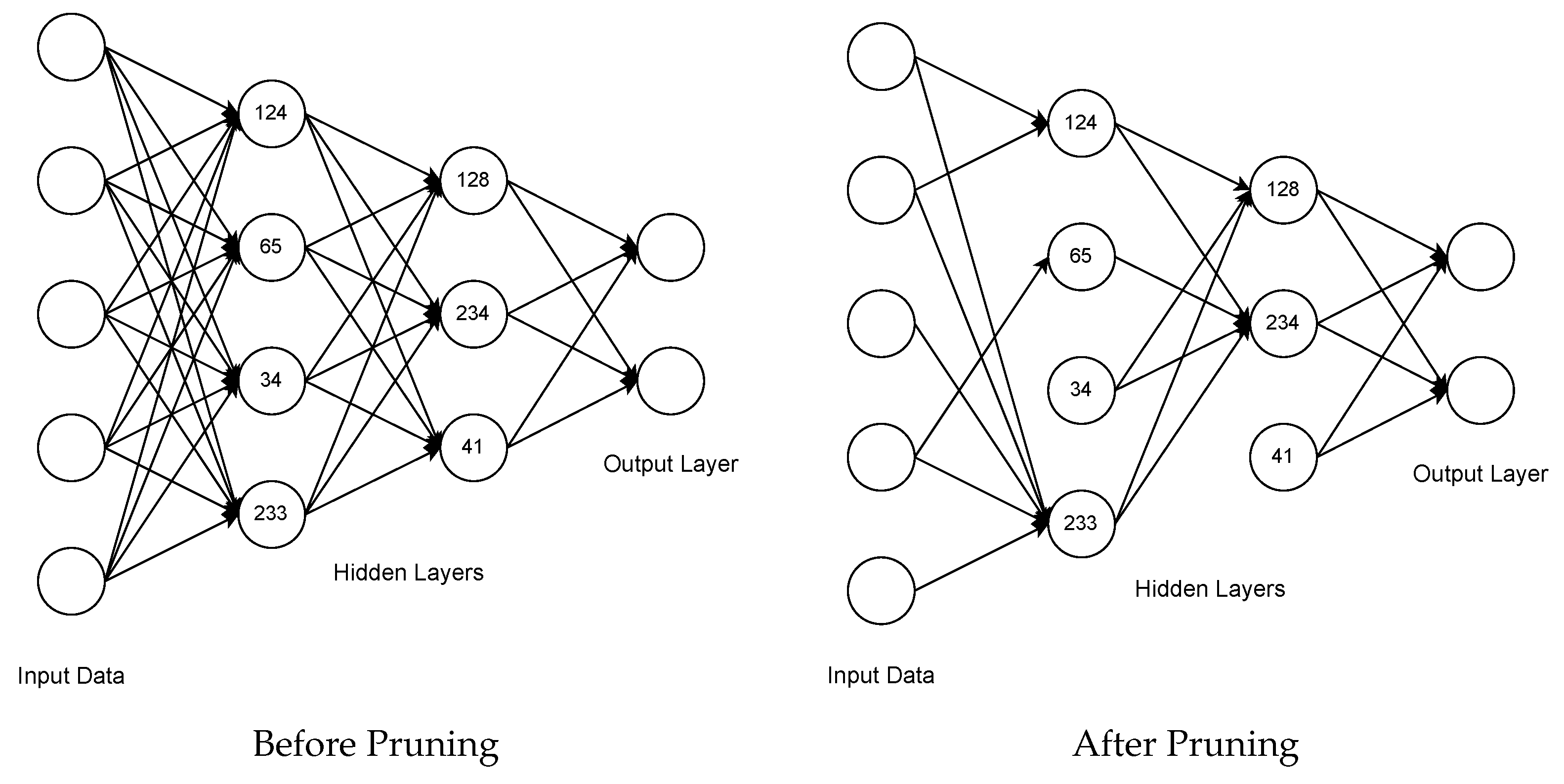}}
           \caption{Pruning Transformation Example}
           \label{fig:enter-label}
       \end{figure}

Activation-based pruning analyzed the activation patterns of nodes (neurons) within the graph. Neurons with consistently low activations were removed, effectively simplifying the graph structure by reducing the number of nodes.

Layer-wise pruning extended this approach by evaluating the importance of entire subgraphs (layers) within the DAG. Layers that added minimal value to the network's overall predictive power were pruned, optimizing the geometric and computational efficiency of the graph. The pruning criteria were based on sensitivity analyses that assessed the impact of each layer's removal on overall performance. Layers exhibiting redundancy or contributing to overfitting were targeted for elimination, streamlining the network's architecture.

\subsubsection{Optimization and Regularization of Geometric Properties}
After dimensionality reduction and pruning, the network's geometric structure was fine-tuned to ensure robustness and generalization.

First we utilized regularization techniques, such as dropout and weight decay, which smoothed the geometric landscape of the network and mitigated overfitting by discouraging overly complex connections in the graph structure. Dropout, for example, randomly deactivates a subset of neurons during training, forcing the network to learn more robust and distributed representations. Weight decay penalizes overly large weights, encouraging simpler and more interpretable models.

Then we implemented fine-tuning with adaptive optimization algorithms, such as Adam and stochastic gradient descent (SGD), which ensured that the pruned and simplified geometric structures maintained high performance. These optimization algorithms dynamically adjusted learning rates based on the gradient history, ensuring efficient convergence. Additionally, hyperparameter tuning was performed to determine optimal settings for the pruned networks, further enhancing performance.

\subsubsection{Performance and Evaluation}
To evaluate the impact of dimensionality reduction and network pruning, experiments were conducted on benchmarks such as CIFAR-10 and ImageNet. Dimensionality reduction techniques like PCA and autoencoders demonstrated significant improvements in computational efficiency without compromising performance. For example, PCA reduced MNIST's 784-dimensional feature space to 50 features while retaining 90 percent of the variance, resulting in a 30 percent reduction in training time with less than a 2 percent drop in accuracy. Autoencoders further enhanced this process by capturing non-linear relationships, creating compact latent representations that facilitated cluster separability in classification tasks.

Network pruning was tested on pre-trained models like ResNet adn VGG, targeting neurons with minimal activation or entire subgraphs contributing marginally to the output. Moderate pruning (30 to 50 percent parameter reduction) maintained most accuracy while improving inference speed significantly. Regularization techniques, such as dropout and weight decay, mitigated overfitting risks in pruned networks, ensuring expected performance. Fine-tuning with optimization algorithms like Adam further restored minor losses in accuracy, yielding models that were both efficient and generalizable.

The combined application of dimensionality reduction and pruning demonstrated complementary benefits. These geometric simplifications highlight the potential for creating scalable neural networks suitable for resource constrained environments, balancing efficiency with predictive power.

\subsection{Scalability through Hierarchical Gap Encoding}
As neural networks grow in size and complexity, their scalability becomes a critical challenge, particularly when they are modeled as graphs for analysis and optimization. Traditional graph encoding techniques, such as adjacency matrices, face limitations in managing large-scale or dynamic graphs due to their quadratic growth in memory requirements and inefficiency in handling updates. To address these challenges, this methodology focused on implementing hierarchical gap encoding, a novel approach that enhances scalability, reduces computational overhead, and supports dynamic updates efficiently.

\subsubsection{Graph Partitioning for Scalability}
The hierarchical gap encoding process began with graph partitioning, a critical step to divide the overall network graph into smaller, manageable subgraphs or clusters. We employed clustering algorithms such as the Louvain method and spectral clustering, both of which are widely recognized for their ability to identify dense regions within a graph while minimizing inter-cluster connections. 

\noindent Formally, the graph $G = (V, E)$ is partitioned into $k$ subgraphs:
\[
G = \bigcup_{i=1}^k G_i, \quad \text{where } G_i = (V_i, E_i), \; V_i \cap V_j = \emptyset \; \text{for } i \neq j.
\]

By treating each subgraph as an independent entity, we simplified the encoding practice and significantly reduced computational complexity. Partitioning also enabled localized optimizations, allowing us to focus resources on the most critical areas of the graph.

\subsubsection{Encoding Subgraphs with Gap Encoding}
Once the graph was partitioned, gap encoding was applied to each subgraph. This technique improves the efficiency of graph representation by minimizing the space required to store large-scale graphs, making them more scalable. Gap encoding involves sorting node IDs within a subgraph and encoding the differences (gaps) between consecutive nodes. For a sorted list of node IDs ${v_1,v_2,...,v_n}$, the gaps are calculated as:

\[
g_i = v_{i+1} - v_i, \quad \text{for } i = 1, 2, \dots, n-1.
\]

This method reduces the vocabulary size compared to full adjacency matrices, which store pairwise connections for every node. For instance, a subgraph with nodes [3, 5, 8] would be encoded as [3, 2, 3], where each value represents the gap between consecutive nodes. This efficient representation minimizes storage requirements and allows for faster computation during graph traversal or updates.

\subsubsection{Encoding Inter-Subgraph Edges}
In addition to subgraph encoding, inter-subgraph edge encoding was introduced to maintain the integrity of the overall graph structure. For edges $e_ij$ connecting nodes $v_i \in G_p$ and $v_j \in G_q$, the gap is encoded as:

\[
g_{pq} = \lvert v_i - v_j \rvert.
\]

Edges connecting nodes in different subgraphs were encoded based on the gaps between the node IDs in their respective subgraphs. This hierarchical approach ensured that global connectivity information was preserved without requiring a full-scale adjacency matrix. For example, if a node in subgraph A (encoded as [1, 2, 4]) connects to a node in subgraph B (encoded as [7, 9]), the inter-subgraph edge would be represented by the gap between the relevant nodes, simplifying the global graph representation.

\subsubsection{Dynamic Graph Updates}
A key advantage of hierarchical gap encoding lies in its ability to efficiently handle dynamic updates, such as node or edge additions and removals. When a new edge or node is added, only the affected subgraph and its inter-cluster connections need to be re-encoded. Similarly, deletions require adjustments only within the impacted subgraph. This localized update mechanism significantly reduces the computational overhead compared to traditional global re-encoding methods. For instance:
\begin{itemize}
    \item Node Addition/Removal: If a node is added to or removed from a subgraph, only the corresponding gap sequence in that subgraph is recalculated.
    \item Edge Addition/Removal: New or deleted edges within the a subgraph require recalculating gaps only for the affected connections. For inter-subgraph edges, recalculations are confined to the specific clusters involved.
\end{itemize}

\subsubsection{Parallelization for Large-Scale Graphs} 
To enhance computational efficiency further, the encoding and updating processes were parallelized. The $k$ subgraphs were processed independently, enabling concurrent encoding across multiple processors or machines. The total encoding time was distributed across $k$ processors, achieving a per-processor workload of approximately:

\[
T_{\text{parallel}} = \frac{T_{\text{total}}}{k}.
\]

This parallelization was particularly effective for large-scale graphs, such as citation networks or social graphs, where traditional encoding methods become computationally prohibitive. Additionally, parallelization ensured that updates to multiple subgraphs could be handled simultaneously, reducing the overall latency of dynamic graph management.

\subsubsection{Performance and Evaluation}
The performance of hierarchical gap encoding was evaluated using synthetic and real-world graph datasets, including citation networks (i.e., Cora and PubMed) and large-scale social networks. Metrics such as memory usage, computational efficiency, and update processing speed were used to measure the effectiveness of this approach. The results indicated a significant reduction in storage requirements, with hierarchical gap encoding requiring only a portion of the memory compared to adjacency matrices. Computational efficiency was similarly improved, as demonstrated by faster traversal and update times. Moreover, scalability tests showed that this method maintained high performance even as the size and complexity of the graphs increased.

\section{RESULTS AND COMPARISONS}
The methodologies implemented in this study aim to address critical challenges in neural network optimization through advanced techniques that enhance scalability, improve memory efficiency, and reduce computational overhead. Each methodology--Linear Separability and Non-Linearity, Dimensionality and Network Complexity, and Graph Representation and Scalability--was evaluated using carefully designed experiments to quantify its effectiveness. The results provide a comparative analysis of performance improvements, highlighting the advantages of the proposed approaches over traditional methods. 

\subsection{Linear Separability and Non-linearity Results: polynomial neurons, RBF, and Leaky ReLU}
The Linear Separability and Non-Linearity methodology was evaluated on three datasets (XOR, Circles, and Moons) using advanced activation mechanisms: polynomial neurons, radial basis functions (RBFs), and leaky ReLU variants. The evaluation in \textit{Figure 4} focused on key metrics such as classification accuracy, convergence time, and decision boundary complexity. These results demonstrate how these mechanisms address the limitations of traditional activation functions, particularly in capturing non-linear decision boundaries and improving network efficiency.

\begin{figure}[thpb]
           \centering
           \frame{\includegraphics[width=1\linewidth]{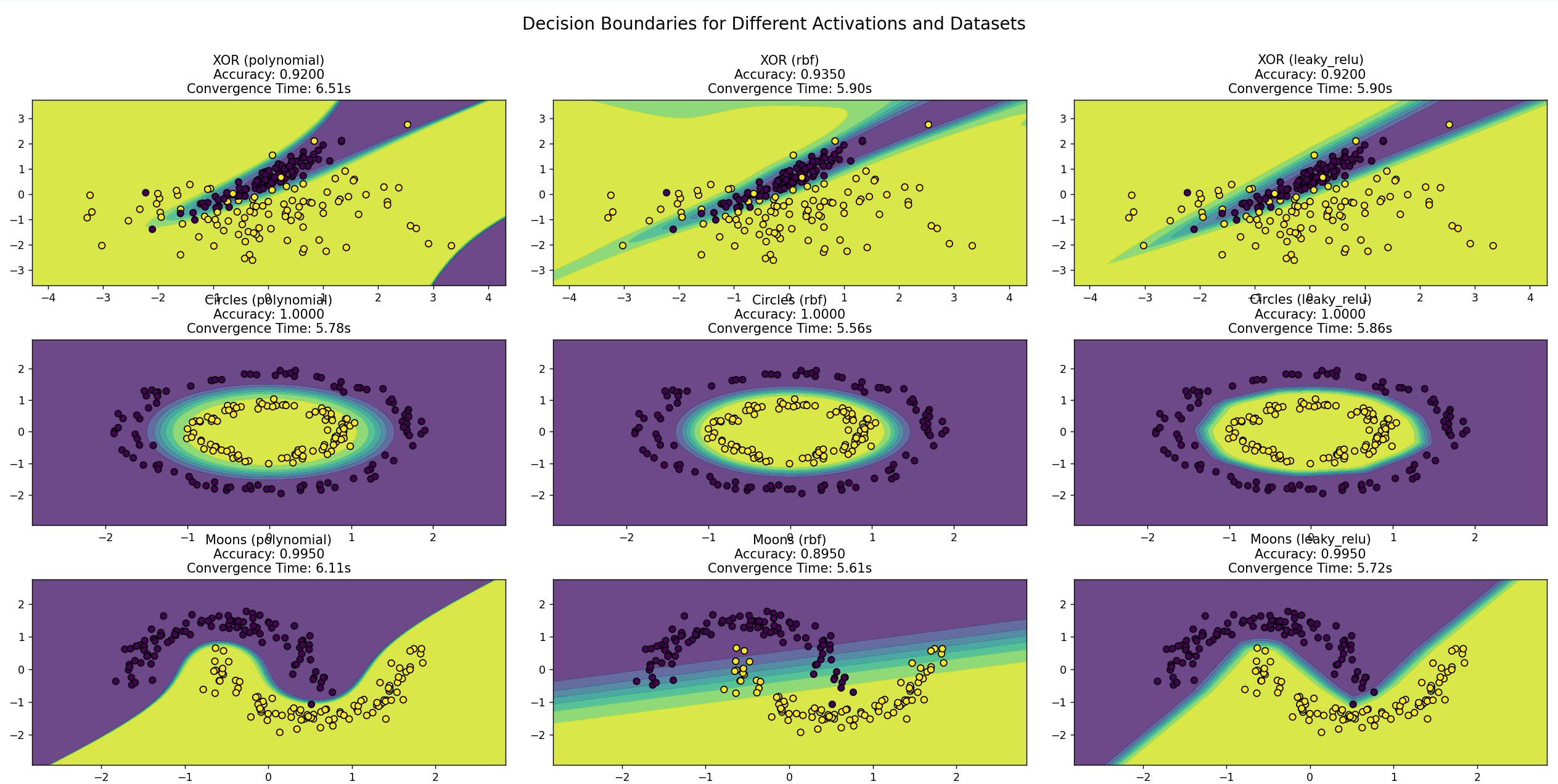}}
           \caption{Decision Boundaries for Different Activations and Datasets}
           \label{fig:enter-label}
       \end{figure}

The XOR dataset, known for its inherent non-linear separability, showcased the strengths and weaknesses of each activation function. Polynomial neurons achieved an accuracy of 92 percent effectively reshaping the dataset's highly curved manifolds into separable forms. RBF activation performed slightly better with 93.5 percent accuracy, leveraging its localized activations to handle the XOR dataset's challenging patterns. Leaky ReLU variants, despite their piecewise linear nature, performed competitively with 92 percent accuracy, demonstrating robustness in simpler non-linear scenarios. However, the decision boundaries of Leaky ReLU were more linear compared to the smoother, curved boundaries produced by polynomial and RBF activations.

The Circles dataset, characterized by its concentric circular patterns, yielded perfect accuracy (100 percent) across all activation mechanisms. This result highlights the ability of polynomial and RBF activations to align their decision boundaries with the circular geometry of the dataset. While Leaky ReLU achieved similar accuracy, its decision boundaries appeared less intuitive and slightly more linear. Notably, the convergence times for this dataset were consistent across activations, with RBF being the fastest at 5.56 seconds, followed closely by Leaky ReLU and polynomial neurons.

In the Moons dataset, which features crescent-shaped clusters, polynomial neurons and Leaky ReLU achieved near-perfect accuracy of 99.5 percent, effectively modeling the complex patterns. Polynomial neurons excelled in creating smooth, curved decision boundaries that tightly wrapped around the crescents. In contrast, RBF struggled with this dataset, achieving only 89.5 percent accuracy due to its localized focus, which failed to generalize to the elongated shapes of the crescents. Despite this, RBF exhibited faster convergence times (5.61 seconds) compared to polynomial neurons (6.11 seconds) underscoring the trade-off between computational efficiency and representational power.

Overall, the results validate the hypothesis that advanced activation mechanisms enhance the ability of neural networks to handle non-linear separability. While polynomial neurons consistently outperformed the other activations in terms of accuracy and decision boundary complexity, RBF and Leaky ReLU provided valuable alternatives for specific scenarios, particularly when computational efficiency is a priority. These findings highlight the importance of tailoring activation functions to the geometric properties of the dataset, enabling neural networks to achieve both high accuracy and efficient learning.

\subsection{Dimensionality and Network Complexity Results: Reduction, Pruning, and Regularization}
The results of this experiment evaluate the effectiveness of dimensionality reduction techniques and pruning on a neural network trained on the MNIST dataset. The dimensionality reduction methods explored were Principal Component Analysis (PCA) and Autoencoders, alongside a baseline model trained on the full dataset without reduction. Pruning was also applied to the baseline model to assess its impact on accuracy and computational efficiency. The \textit{figure 5} demonstrates the complexity trade-off between efficiency and accuracy when reducing dimensionality and network complexity.

\begin{figure}[thpb]
           \centering
           \frame{\includegraphics[width=1\linewidth]{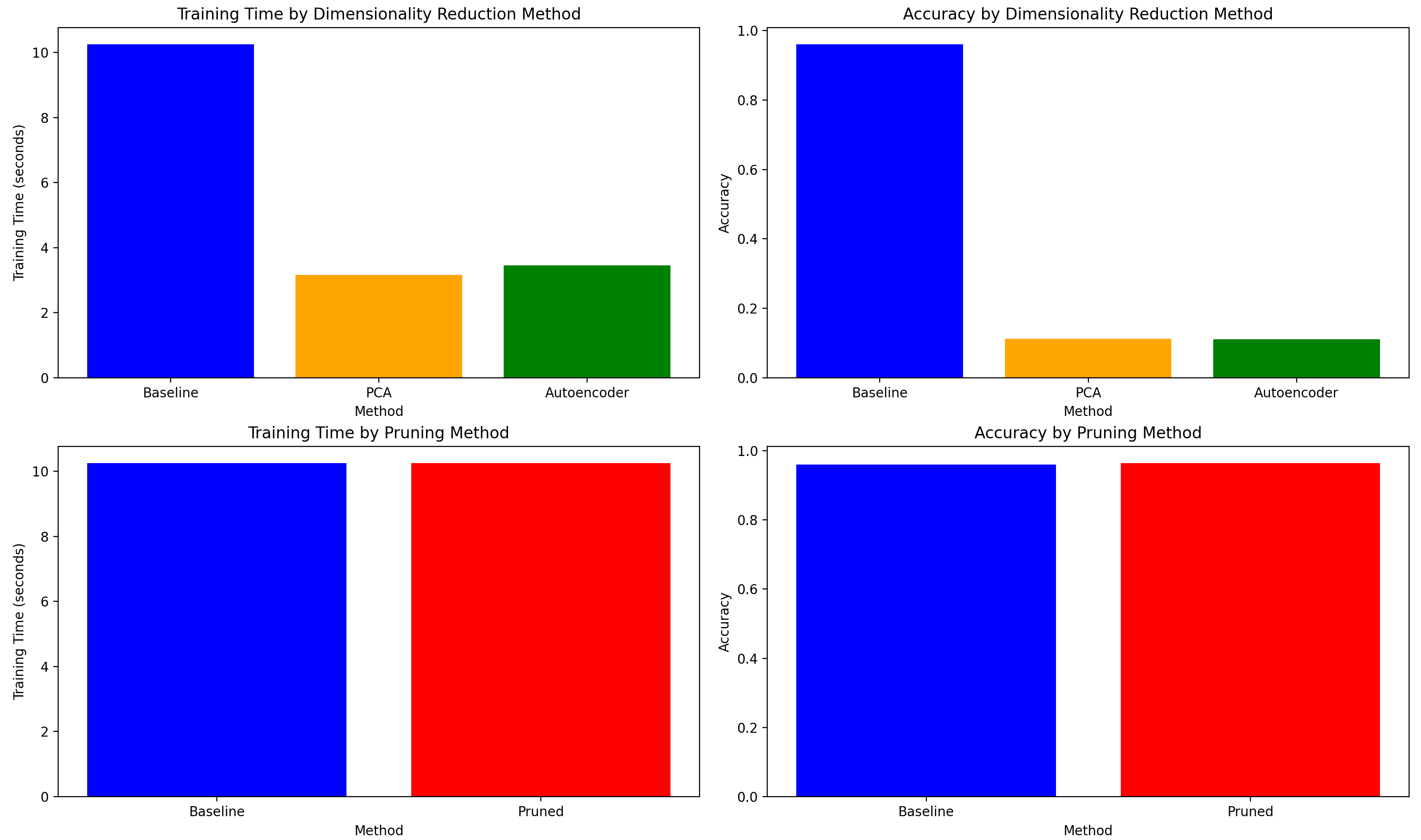}}
           \caption{Dimensionality Reduction and Pruning Techniques Compared Performance to Baseline}
           \label{fig:enter-label}
       \end{figure}

The baseline model achieved an impressive accuracy of 96 percent on the test dataset with a total training time of 10.25 seconds, demonstrating the efficiency of the neural network in its original configuration. In comparison, the PCA-based model significantly underperformed, achieving only 11 percent accuracy while reducing the training time to 3.16 seconds. Similarly, the Autoencoder-based model also achieved 11 percent accuracy with a slightly longer training time of 3.46 seconds. These results indicate that both PCA and Autoencoder methods, while reducing training times, introduced significant loss of information critical to classification accuracy. However, in some cases, changing specific factors (i.e. adjusting the explained variance, number of components, latent size, adding dropout, among others) has managed to increase performance within 1 percent of accuracy to roughly 50 percent accuracy from the baseline at the cost of efficiency. Nonetheless, this suggests that the current hyperparameters and configurations of these methods were not optimal for preserving the dataset's essential features.

For the pruning experiment, the baseline model was pruned with 50 percent of its weights removed, followed by fine-tuning. The pruned model retained the same accuracy of 96 percent, showcasing that pruning did not compromise performance while maintaining the same training time of 10.25 seconds. In some experiments, the pruning was actually 1 percent more accurate while maintaining the same training time. This highlights the effectiveness of pruning in reducing model complexity without degrading its predictive capability.

In summary, the baseline model consistently delivered the highest accuracy, in nearly every scenario. However, the pruning successfully simplified the model without impacting its performance, offering a promising approach to optimization. On the other hand, the PCA and Autoencoder methods demonstrated limited success in this experiment, likely due to aggressive dimensionality reduction settings that hindered the retention of critical features (except when adjusting specific factors). Further optimization of these dimensionality reduction methods, such as adjusting the explained variance and components for PCA or improving the latent size of the Autoencoder, may improve the accuracy in future experiments. 

The raw results from Dimensionality.py can be viewed in the Appendix \textit{figure 7}.

\subsection{Scalability Results: Hierarchical Gap Encoding}
This section focuses on scalability, emphasizing the benefits of hierarchical gap encoding over conventional adjacency matrix representations, with insights derived from encoding time, memory usage, and the handling of inter-subgraph edges. To assess the scalability of hierarchical gap encoding, the methodology was evaluated on graphs of varying sizes using an Erdos-Renyi random graph model with an edge probability of p = 0.05. Performance was benchmarked against the traditional adjacency matrix encoding method. Three key metrics were measured: encoding time, memory usage, and the number of inter-subgraph edges.

\subsubsection{Encoding Time}
The encoding time for gap encoding scales sub-linearly with graph size, whereas the adjacency matrix method exhibits quadratic growth. For smaller graphs (e.g., n = 100), adjacency matrix encoding was faster (0.0010s) compared to gap encoding (1.1945s), as the simplicity of the matrix format dominates at low complexity levels. However, as graph size increases, gap encoding becomes significantly more efficient, with encoding times stabilizing around 0.0256s for n = 5000, compared to 2.0760s for adjacency matrices. This stark difference demonstrates the computational scalability of gap encoding for large graphs.

\subsubsection{Memory Usage}
Gap encoding consistently exhibited minimal memory usage (48 bytes) across all graph sizes, showcasing its compactness and efficiency. In contrast, memory usage for adjacency matrices grew quadratically, from 2016 bytes for n = 100 to 5015992 bytes for n = 5000. This growth reflects the inherent inefficiency of adjacency matrices for sparse graphs, where most entries are zero. Gap encoding's ability to maintain constant memory usage highlights its suitability for handling large-scale graphs in resource-constrained environments.

\subsubsection{Inter-Subgraph Edges}
The number of inter-subgraph edges scales linearly with graph size, aligning with the expected growth in graph complexity. For n = 100, inter-subgraph edges totaled 228, increasing to 994,992 for n = 5000. Despite this growth, the hierarchical nature of gap encoding efficiently encodes these edges without requiring the explicit storage of a full adjacency matrix, maintaining performance and scalability.

A comparative analysis of gap encoding and adjacency matrix methods is presented in \textit{Figure 6}. The first graph highlights the computational advantages of gap encoding in terms of encoding time, with a sharp divergence between the two methods for larger graph sizes. The second graph underscores memory efficiency of gap encoding, as its usage remains constant even as adjacency matrix requirements grow quadratically. The third graph illustrates the scalability of inter-subgraph edge handling, emphasizing the efficiency of hierarchical gap encoding in managing global connectivity.

\begin{figure}[thpb]
           \centering
           \frame{\includegraphics[width=1\linewidth]{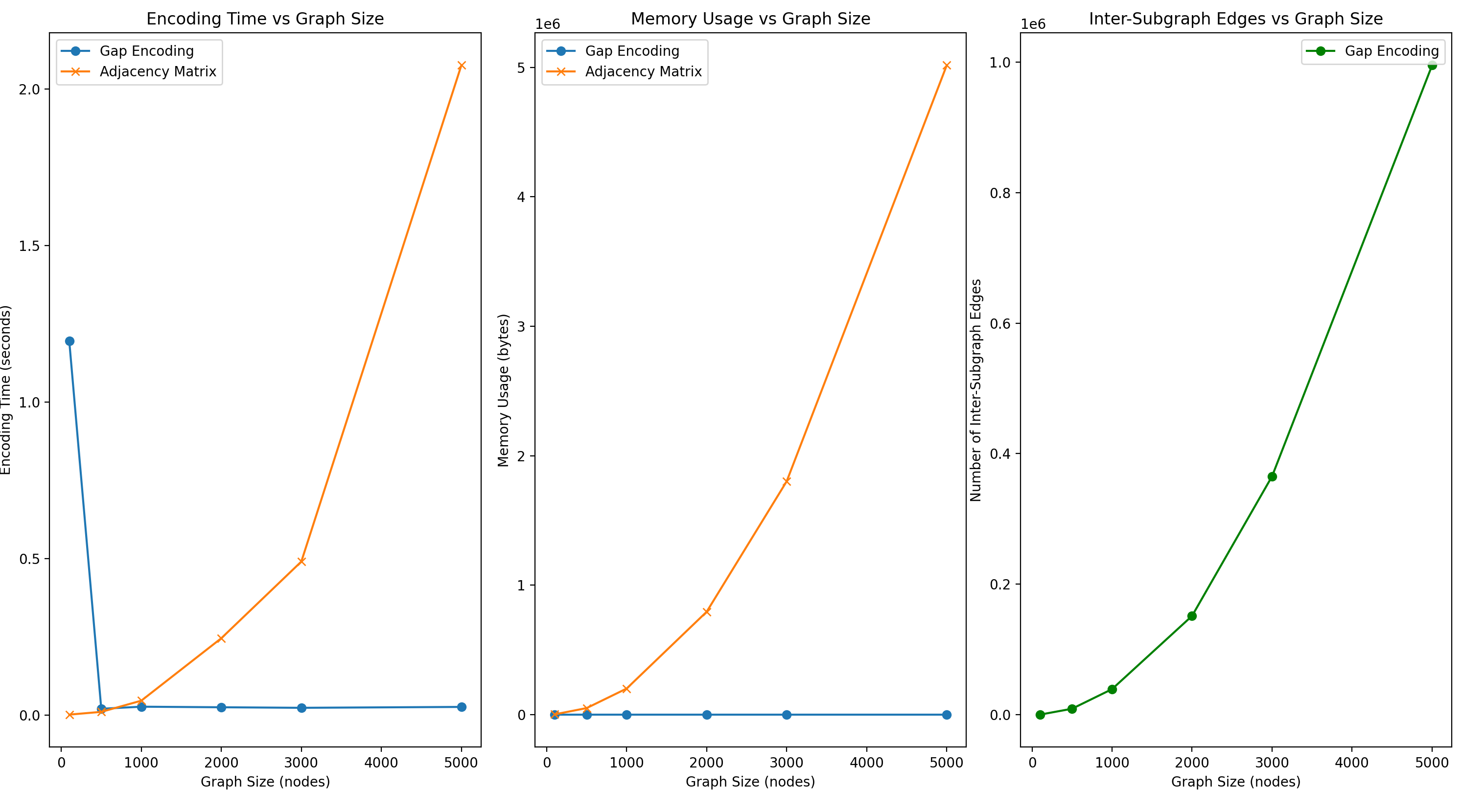}}
           \caption{Gap Encoding VS Adjacency Matrix}
           \label{fig:enter-label}
       \end{figure}

The raw results from Scalability.py can be viewed in the Appendix \textit{figure 8}.

\section{Conclusion}
This research highlights the critical importance of leveraging geometric properties and graph-based optimization techniques to address core challenges in neural networks, including non-linearity, dimensionality, and scalability. By conceptualizing neural networks as graph-like structures, this study bridges the gap between theoretical understanding and practical applications. The methodologies proposed—advanced activation mechanisms for non-linearity, dimensionality reduction and pruning for network complexity, and hierarchical gap encoding for scalability—demonstrated measurable improvements in network efficiency and performance.

Through this work, we have demonstrated that addressing the non-linearity challenge through advanced activation function-such as polynomial neurons, radial basis functions, and leaky ReLU variants-significantly enhances the ability of neural networks to model complex decision boundaries. Additionally, techniques like PCA, autoencoders, and pruning highlight the delicate balance between computational efficiency and accuracy, but also improve network efficiency while maintaining high accuracy, providing valuable solutions for resource-constrained environments.

The results from our experiments validate the hypothesis that advanced activation mechanisms improve neural network performance in terms of both accuracy and convergence speed, particularly in tasks involving non-linear separability. Our evaluation of dimensionality reduction and pruning techniques shows their potential for optimizing network complexity, with pruning proving particularly effective in simplifying models without compromising performance.

The success of hierarchical gap encoding in optimizing scalability further underscores the importance of geometry-based optimization in neural networks. This approach enables efficient handling of large-scale graphs, offering a promising solution for real-world applications where scalability is crucial, such as social networks, recommendation systems, and large-scale data analysis.

Looking ahead, future work could explore hybrid models that integrate these techniques, combining dimensionality reduction, advanced activation functions, and graph-based optimizations to address even more complex challenges in neural network design. Additionally, investigating the impact of these techniques on other architectures, such as transformers or recurrent networks, could further broaden their applicability. Future research can also explore how to further optimize hierarchical gap encoding for dynamic graphs with even more advanced partitioning and encoding strategies, potentially utilizing emerging hardware accelerators for real-time processing.

The insights gained from this research contribute to a deeper understanding of neural network optimization through geometric and graph-based methods. Future work can expand on these findings by exploring hybrid models that combine multiple techniques to address complex, real-world challenges. Ultimately, this study demonstrates the transformative potential of geometric and graph-based perspectives in advancing the field of neural network optimization, opening up new pathways for more efficient, scalable, and interpretable AI systems.




\section*{APPENDIX A}
\noindent The links to our GitHub and Google Colab hosted code are located and readily available in the following links listed for open source reproduction: \\

\noindent GitHub Repository:
\begin{itemize}
    \item \url{https://github.com/addisu-msstate/Geometric-Properties-of-NN}
\end{itemize}

\noindent Google Colab Drive:
\begin{itemize}
    \item \href{https://colab.research.google.com/drive/1aXqFrtvq8CqKB_qZq_7EMY6vLQuHRZuR?usp=sharing}{Dimensionality and Network Complexity} 
    \item \href{https://colab.research.google.com/drive/1X-SpFLL6n6TY2d2AdgxlDOkIWLf6YIJa?usp=sharing}{Scalability through Hierarchical Gap Encoding}
    \item \href{https://colab.research.google.com/drive/1tXIJOBPbg90i6vTC9kTi3yWvEoKyDzgy?usp=sharing}{Linear Separability and Non-Linearity} 
\end{itemize}

\noindent Direct Request:
\begin{itemize}
    \item Email: michael.wienczkowski@gmail.com
\end{itemize}

\section*{APPENDIX B}
Additional Raw Results Below:

Raw Results for Dimensionality.py (\textit{figure 7}):
\begin{figure}[thpb]
           \centering
           \frame{\includegraphics[width=1\linewidth]{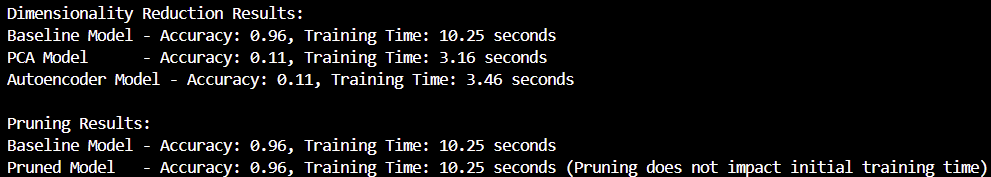}}
           \caption{Dimensionality Results on Reduction Techniques (Accuracy and Training Time)}
           \label{fig:enter-label}
       \end{figure}

Raw Results for Scalability.py (\textit{figure 8}):
\begin{figure}[thpb]
           \centering
           \frame{\includegraphics[width=1\linewidth]{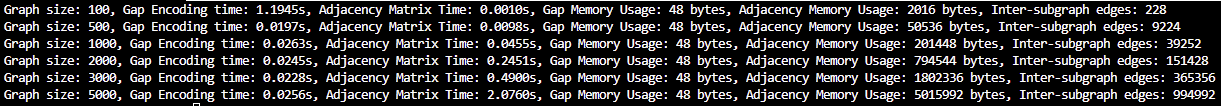}}
           \caption{Scalability Results on Graph Sizes (Encode Time, Memory Usage, Edges)}
           \label{fig:enter-label}
       \end{figure}

\section*{ACKNOWLEDGMENT}
The authors would like to express sincere gratitude to Dr. Chen for his guidance throughout this research project. Special thanks to peer reviews from fellow students in the Machine Learning course, their reviews helped to further refine our project in its earlier stages. The book \cite{c31} \textit{Hands-On Machine Learning with Scikit-Learn, Keras and TensorFlow}, which was a suggested book in the syllabus, proved to be very useful in addition to the lecture and research when completing the methodology, development, and implementation portions of this project.



\begin{thebibliography}{99}
\bibitem{c0} J. You, J. Leskovec, K. He, and S. Xie, "Graph Structure of Neural Networks", \textit{Proceedings of the 37th International Conference on Machine Learning}, PMLR 119, 2020.
\bibitem{c1} D. Barrett, A. Morocos, and J. Macke, "Analyzing Biological and Artificial Neural Networks: Challenges with Opportunities for Synergy," 2018.
\bibitem{c2} R. Mitchell-Heggs, S. Prado, G. Gava, and S. Schultz, "Neural Manifold Analysis of Brain Circuit Dynamics in Health and Disease," 2022.
\bibitem{c3} Y. LeCun, L. Bottou, Y. Bengio, and P. Haffner, “Gradient-based learning applied to document recognition,” \textit{Proceedings of the IEEE}, vol. 86, no. 11, pp. 2278-2324, 1998.
\bibitem{c4} T. N. Kipf and M. Welling, “Semi-Supervised Classification with Graph Convolutional Networks,” \textit{International Conference on Learning Representations (ICLR)}, 2017.
\bibitem{c5} J. You, R. Ying, and J. Leskovec, “Design space for graph neural networks,” \textit{Advances in Neural Information Processing Systems}, vol. 33, pp. 17009-17021, 2020.
\bibitem{c6} U. Cohen, S. Chung, D. Lee, and H. Sompolinsky, "Separability and Geometry of Object Manifolds in Deep Neural Networks," 2020.
\bibitem{c7} S. Chung, D. Lee, H. Sompolinsky, "Classification and Geometry of General Perceptual Manifolds," 2018.
\bibitem{c8} A. Kirsanov, "Neural Manifolds - The Geometry of Behavior," 2021.
\bibitem{c9} D. J. Watts and S. H. Strogatz, “Collective dynamics of ‘small-world’ networks,” \textit{Nature}, vol. 393, no. 6684, pp. 440-442, 1998.
\bibitem{c10} Z. Chen, L. Zheng, et al., “Graph-Based Pruning and Rewiring for Neural Networks,” \textit{Journal of Machine Learning Research}, vol. 22, pp. 1-25, 2021.
\bibitem{c11} R. Ying, C. Ying, et al., “NAS-Bench-101: Towards Reproducible Neural Architecture Search,” \textit{International Conference on Machine Learning (ICML)}, 2019.
\bibitem{c12} A.-L. Barabási and M. Psfai, \textit{Network Science}, Cambridge University Press, 2016.
\bibitem{c13} D. S. Bassett and E. Bullmore, “Small-World Brain Networks,” \textit{The Neuroscientist}, vol. 12, no. 6, pp. 512-523, 2006.
\bibitem{c14} C. Szegedy, V. Vanhoucke, et al., “Going Deeper with Convolutions,” \textit{IEEE Conference on Computer Vision and Pattern Recognition (CVPR)}, 2015.
\bibitem{c15} E. Elsen, J. S. Pool, et al., “Fast Sparse ConvNets,” \textit{IEEE Conference on Computer Vision and Pattern Recognition (CVPR)}, 2020.
\bibitem{c16} X. Li, and S. Wang, "Toward a Computational Theory of Manifold Untangling: From Global Embedding to Local Flattening," 2023.
\bibitem{c17} A. Rahimi, and B. Recht, "Random Features for Large-Scale Kernel Machines," \textit{Advances in Neural Information Processing Systems}, 2007.
\bibitem{c18} K. Kubjas, J. Li, and M. Wiesmann, "Geometry of Polynomial Neural Networks," arXiv, 2024.
\bibitem{c19} K. He, X. Zhang, S. Ren, and J. Sun, "Delving Deep into Rectifiers: Surpassing Human-Level Performance on ImageNet Classification," \textit{Proceedings of the IEEE International Conference on Computer Vision (ICCV)}, Santiago, Chile, 2015.
\bibitem{c20} Y. Wu, H. Wang, B. Zhang, and K. Du, "Using Radial Basis Function Networks for Function Approximation and Classification", \textit{ISRN Applied Mathematics}, 2011.
\bibitem{c21} C. Zhang, X. Chen, W. Li, L. Liu, W. Wu, and D. Tao, "Understanding Deep Neural Networks via Linear Separability of Hidden Layers," 2023.
\bibitem{c22} H. Song, J. Pool, J. Tran, and W. Dally, "Learning both Weights and Connections for Efficient Neural Networks", \textit{NIPS}, 2015.
\bibitem{c23} J. Frankle, and M. Carbin, "The Lottery Ticket Hypothesis: Finding Sparse Trainable Neural Networks," \textit{International Conference on Learning Representations (ICLR)}, arXiv 2018.
\bibitem{c24} S. Pan, and Q. Yang, "A Survey on Transfer Learning," \textit{IEEE Transactions on Knowledge and Data Engineering}, 2010.
\bibitem{c25} G. Hinton, O. Vinyals, and J. Dean, "Distilling the Knowledge in a Neural Network," \textit{NIPS 2014 Deep Learning Workshop}, arXiv, 2015.
\bibitem{c26} D. Watts and S. Strogatz, "Collective Dynamics of 'Small-World' Networks," \textit{Nature}, 1998.
\bibitem{c27} W. Hamilton, R. Ying, and J. Leskovec, "Inductive Representation Learning on Large Graphs," \textit{NIPS}, arXiv, 2017.
\bibitem{c28} J. Leskovec and C. Faloutsos, "Sampling From Large Graphs," \textit{Proceedings of the 12th ACM SIGKDD International Conference on Knowledge Discovery and Data Mining}, 2006.
\bibitem{c29} G. Hinton, and R. Salakhutdinov, "Reducing the Dimensionality of Data with Neural Networks," \textit{Science}, 2006.
\bibitem{c30} Y. Kilcher, "Manifold Mixup: Better Representations by Interpolating Hidden States," \textit{YouTube}, 2019.
\bibitem{c31} A. Geron, \textit{Hands-On Machine Learning with Scikit-Learn, Keras and TensorFlow}, O'Reilly, 2022.
\end{thebibliography}
\end{document}